\documentclass[10pt,letterpaper]{article}

\usepackage[utf8]{inputenc}
\usepackage[T1]{fontenc}
\usepackage[margin=1in]{geometry}
\usepackage{times}
\usepackage{microtype}
\usepackage{amsmath,amssymb}
\usepackage{graphicx}
\usepackage{booktabs}
\usepackage{makecell}
\usepackage{xcolor}
\usepackage{caption}
\usepackage{subcaption}
\usepackage{enumitem}
\usepackage{needspace}
\usepackage{listings}
\usepackage[numbers,sort&compress]{natbib}
\usepackage{tikz}
\usetikzlibrary{positioning,arrows.meta,calc,fit,backgrounds}
\usepackage[colorlinks=true,allcolors=blue!55!black,breaklinks=true]{hyperref}

\definecolor{pmnavy}{HTML}{0B2A4A}
\definecolor{pmblue}{HTML}{1F6FEB}
\definecolor{pmteal}{HTML}{169F84}
\definecolor{pmred}{HTML}{E8593B}
\definecolor{pmamber}{HTML}{C88612}
\definecolor{pmslate}{HTML}{465768}
\definecolor{pmtint}{HTML}{EAF2FD}

\newcommand{\cmark}{\textcolor{pmteal}{\textbf{\checkmark}}}
\newcommand{\xmark}{\textcolor{pmslate!70}{\ding{55}}}
\usepackage{pifont}
\renewcommand{\xmark}{\textcolor{pmslate!70}{\ding{55}}}
\newcommand{\pmark}{\textcolor{pmamber}{$\circ$}}

\lstdefinestyle{pm}{
  basicstyle=\ttfamily\small,
  breaklines=true,
  frame=single,
  rulecolor=\color{pmblue!40},
  backgroundcolor=\color{pmtint!50},
  numbers=none,
  keywordstyle=\color{pmblue}\bfseries,
  commentstyle=\color{pmslate}\itshape,
  showstringspaces=false,
}
\title{\textbf{PROJECTMEM: A Local-First, Event-Sourced Memory and\\
Judgment Layer for AI Coding Agents}}

\author{
  Ripon Chandra Malo\thanks{Corresponding author: \texttt{ripon.malo@utah.edu}} \\
  University of Utah
  \and
  Tong Qiu \\
  University of Utah
}
\date{June 2026; revised September 2026}

\begin{document}
\maketitle

\begin{abstract}
\noindent
AI coding agents often lose project-specific rationale across sessions and may repeat approaches
that previously failed. The bottleneck is often not model capability but missing coding agent
memory: a durable record of what was decided, tried, and ruled out. We present projectmem, an
open-source, local-first memory and judgment layer for AI coding agents, which records development
as an append-only, plain-text log of typed events (issues, attempts, fixes, decisions, and notes) and
deterministically projects it into compact summaries served through the Model Context Protocol
(MCP). When invoked before an edit, a file-scoped advisory precheck returns recorded failures, open
issues, and churn for the target file, a design point we call Memory-as-Governance. We evaluate
feasibility through an author-run, six-month self-study on two machines, a controlled latency
benchmark, and client-compatibility tests. The study contains 3{,}228 unique events from 27 projects;
56\% are issue-lifecycle events, and 86 of 427 distinct issues closed by a recorded fix had at least
one failed attempt before the first fix in event-log order. These histories are candidates for later
warnings, not evidence that warnings prevented failures. On a synthetic repository, the median
analysis time for a 1{,}500-event log decreased from 42.0 seconds to 79.3 milliseconds, while
\texttt{git} subprocesses per check decreased from 1{,}500 to two and the tested warning output was
preserved. The results establish the feasibility of persistent project memory for AI coding agents
and motivate a controlled repeat-failure benchmark for measuring whether coding-agent memory
prevents repeated failures. Source code is available at
\url{https://github.com/riponcm/projectmem}.
\end{abstract}

\vspace{0.4em}

% ═══════════════════════════════════════════════════════════
\section{Introduction}
\label{sec:intro}

Coding agents based on large language models (LLMs) have rapidly become everyday development infrastructure:
from an individual developer rapidly prototyping an idea, to an engineering team shipping features, to a
researcher writing analysis code, developers increasingly drive their work through an AI assistant
rather than typing it by hand. These agents can be powerful within a single session but often lack
durable project-specific state across sessions. When the conversation window closes, the agent
loses access to project-specific rationale that was held only in the conversation. The next session
often begins by re-reading source files, re-asking questions answered
yesterday, re-deriving architectural decisions, and, in a particularly costly failure mode, re-attempting
fixes that have already been tried and have already failed.

This is not only a model-quality problem; it is also an architectural one. Recent empirical work on
agentic software engineering documents exactly these pathologies. An empirical study of failed
agentic pull requests on GitHub finds that they frequently exhibit ``repeated application of the
same fix without proper testing or evolution''~\citep{wheredoagentsfail2026}; a complementary
analysis shows that a single root-cause error propagates through an agent's subsequent decisions
into cascading task failure, and that agents can learn from such failures, but only
after the fact~\citep{zhu2025wherefail}. Surveys of agentic programming likewise identify
cross-session memory and context tracking ``beyond the token limit'' as a central open
challenge~\citep{aiagenticprogramming2025}. The context cost is concrete: re-establishing context
by re-reading a project consumes thousands of tokens per session, and the per-turn cost of
long-context re-submission grows with horizon even under aggressive prompt caching, whereas a
memory system's read cost is roughly fixed after a one-time write~\citep{beyondcontextwindow2026}.
These studies point to a common failure mode: repeated mistakes are costly, and the most useful
moment to surface a prior failure is before the next attempt, not after it.

A natural response is to give agents memory, and a substantial literature now does
(Section~\ref{sec:related}). However, many representative designs share a set of properties poorly suited
to everyday software development: they are built around vector databases or LLM-in-the-loop fact
extraction (introducing nondeterminism and recurring read cost), they are oriented toward
conversational personalization rather than engineering correctness, they are frequently
cloud- or server-hosted (a barrier for sensitive or proprietary code), and, most
importantly, they primarily answer the agent. Even the closest coding-agent peer, which
independently converges on a plain-text, file-based, no-vector-database design for the same reason
(agents ``lose coherence across sessions, forget project conventions, and repeat known
mistakes''~\citep{vasilopoulos2026codified}), supplies project knowledge passively, as
conventions the agent must choose to read. Among the systems we reviewed, none consults the
project's own recorded failures at the point of an action.

\paragraph{Contributions.} This paper is a system and feasibility report. We present
\textbf{projectmem} and make four contributions:
\begin{enumerate}[leftmargin=1.4em,itemsep=2pt]
  \item \textbf{An event-sourced, plain-text memory substrate} for coding agents: an append-only
        log of typed events (issue / attempt / fix / decision / note) from which a compact,
        AI-readable summary is deterministically projected, and in which a decision can
        explicitly retire an earlier one. The log is grep-able, diff-able, and git-native, with no
        vector database and no embeddings (Section~\ref{sec:design}).
  \item \textbf{An advisory judgment layer}: a deterministic, history-derived precheck that, when
        invoked before an edit, surfaces the failed attempts, open issues, and churn recorded for
        the target file, with a git pre-commit hook as a later fallback. We argue that this is an
        underexplored design point, which we call Memory-as-Governance: memory that is consulted
        at the point of action rather than only when the agent chooses to read it
        (Sections~\ref{sec:design}, \ref{sec:related}).
  \item \textbf{A local-first, tool-agnostic implementation}: a native MCP server (17 tools) that
        serves identical memory to MCP-capable clients, bridge files for tools that read
        instructions from the repository, and default-on secret redaction, with no network request
        unless the user enables an update check (Section~\ref{sec:impl}).
  \item \textbf{A feasibility evaluation}: a six-month self-study of 3{,}228 events across 27 projects on two
        machines; a controlled
        latency measurement of the precheck on a synthetic repository; a field observation of how
        the supersession relation was used; and an acceptance suite across three MCP clients and
        Windows. The evaluation shows that the substrate accumulates structured memory and that
        the precheck is cheap enough to call before every edit in the measured synthetic workloads. It
        does not measure how often
        repeated failures are prevented, for which we propose a benchmark
        (Sections~\ref{sec:eval}, \ref{sec:limits}).
\end{enumerate}

% ═══════════════════════════════════════════════════════════
\section{System Design}
\label{sec:design}

\paragraph{Design principles.} projectmem follows four principles: (i) append-only memory: memory
is an event log that the application only appends to and never edits in place, yielding a
replayable audit trail (the log is an ordinary text file, not a cryptographically protected one, so
it can still be edited by hand); (ii) human-legibility: memory is plain text (JSON Lines and
Markdown), so it is grep-able, diff-able, reviewable in a pull request, and versioned by git;
(iii) locality: all state lives in the repository and on the machine, and projectmem makes no
network request unless the user explicitly enables its update check; and (iv) determinism: the
summary the agent reads is a deterministic projection of the log, and the judgment gate is a
deterministic lookup, not a model call.

\paragraph{Event schema.} Development is recorded as typed events (Table~\ref{tab:schema}). Every
event carries a random identifier, its type, a free-text summary, and an ISO-8601 UTC timestamp,
and may carry a location (a path, optionally with a line number), the current git commit, and, for
events captured by a git hook, the capture source and the commit message.

\begin{table}[t]
\centering
\caption{Event types and their type-specific fields. Common fields are listed in the text.}
\label{tab:schema}
\small
\begin{tabular}{lll}
\toprule
Type & Records & Type-specific fields \\
\midrule
\texttt{issue}    & a problem is opened                        & \texttt{issue\_id} (assigned) \\
\texttt{attempt}  & a fix is tried against the active issue    & \texttt{issue\_id}; \texttt{outcome} $\in$ \{\texttt{worked}, \texttt{failed}, \texttt{partial}\} \\
\texttt{fix}      & a confirmed resolution; closes the issue   & \texttt{issue\_id} \\
\texttt{decision} & an architectural or product choice         & \texttt{supersedes} (optional) \\
\texttt{note}     & a durable gotcha or setup detail           & none \\
\bottomrule
\end{tabular}
\end{table}

\paragraph{Projection.} The file the agent reads, \texttt{summary.md}, is regenerated after every
write by a fold over the log, and can be rebuilt at any time with \texttt{pjm regenerate}. It is a
deterministic function of the log and of one hand-edited input, the project-purpose section of
\texttt{PROJECT\_MAP.md}. The fold applies fixed rules: every issue is listed, newest first, as open
or done together with its fix and its three most recent failed or partial attempts; every decision
that has not been retired is listed; only the ten most recent notes are kept; and up to twenty files
cited by events are listed as key files. Because the summary is derived rather than authoritative, it
can always be regenerated from the log; after an interrupted write it may lag until the next
regeneration repairs it (see Durability, below). This contrasts with mutable-memory designs in which
the agent overwrites its own state. A second, hand- and agent-edited file, \texttt{PROJECT\_MAP.md},
describes the repository's structure and stack.

\paragraph{Intent.} Plans and ideas, what the team means to do rather than what has happened, are
kept in a separate file, \texttt{plan.md}, which the developer or the agent edits directly and which
an agent reads through \texttt{get\_plan}. It is deliberately not part of the event log: a plan is
revised as often as it is followed, and recording it as events would mix intentions with outcomes in
a log that is meant to record only what occurred.

\paragraph{Agent protocol.} Because most events are written by agents rather than captured
automatically (Section~\ref{sec:eval}), what the agent is told to do is itself part of the capture
mechanism. The instructions reach the agent on three surfaces: \texttt{AI\_INSTRUCTIONS.md}, returned
by \texttt{get\_instructions}; the bridge files \texttt{CLAUDE.md} and \texttt{AGENTS.md} at the
repository root; and the MCP server's own instructions, which a client may pass to the model when it
connects. All three ask the agent to begin a session by reading the instructions, the summary, and the
project map rather than re-reading the source, and to call the gate before modifying a file. While
working, the agent is asked to log an issue before attempting a fix, to record each attempt separately
with its outcome, to record a fix only once there is evidence that it works, and to record decisions
and durable notes as they arise. On a new project, whose summary and map still hold the placeholder
text written by \texttt{pjm init}, the instructions first direct the agent to populate them from the
repository's manifests and entry points (Setup Mode); once they hold real content, the agent is told
to use them as its starting context and verify current implementation details against source when
correctness depends on them, rather than indiscriminately re-deriving the project (Maintenance
Mode). None of this is
enforced: an agent may skip a read or omit a write, which is why the self-study counts what was
written rather than what should have been.

\paragraph{The judgment gate.} The central design choice is that a logged \texttt{attempt} with
outcome \texttt{failed} is not merely stored for later reading; it is returned to the agent when the
agent is about to work on the same file. The gate, \texttt{precheck\_file(file\_path)}, takes a
path, the event log, and the repository's git history, and returns zero or more warnings, each with
a severity. It reports six things for the path. Four look back 30 days: failed attempts recorded
against the file (the three most recent are listed), high churn (four or more commits touching it),
a recent revert that affected it, and the most recent decision about it. Two are reported
regardless of age: issues located in the file that are still open, and memories citing it that may
be stale because the file has since changed substantially or no longer exists. Three or more failed attempts raise the severity from
warn to block. The check reads only memory and git metadata, never the file's contents; it costs one
pass over the log in memory plus two \texttt{git} invocations, and it is deterministic.

The 30-day window, four-commit churn threshold, and three-failure severity threshold are engineering
defaults; they were not empirically optimized, and this paper does not evaluate their utility.

The gate is file-scoped and advisory. It does not compare a proposed change with an earlier failed
approach; it reports which approaches have already failed on this file, and the agent or developer
decides what to do with that. It is reached in three ways: an agent calls it through MCP before
editing a file, as the project's instructions ask it to; a developer runs \texttt{pjm precheck}; and
a git pre-commit hook runs it after editing but before committing. The installed hook reports and
never blocks a commit; a stricter mode that fails on block-severity warnings is available for manual
or continuous-integration use. A developer can also silence the warnings for a set period
(\texttt{pjm precheck --snooze}); the snooze is itself logged, so the silence stays on the record.

\paragraph{Supersession.} An append-only log raises an immediate question: what happens when a
decision is reversed? Rewriting the earlier event would break the append-only property, and leaving
both in place yields a summary that asserts two incompatible choices. projectmem resolves this with
an explicit relation. A \texttt{decision} event may carry a \texttt{supersedes} field holding the
identifier of the event it replaces. The reference is authored by the actor (the developer through
the CLI or the agent through the \texttt{add\_decision} tool) at the moment of writing, and is never
inferred from the text of either decision. The implementation maintains the following invariants.
\begin{itemize}[leftmargin=1.4em,itemsep=1pt]
  \item The reference is resolved when the decision is written, from a full identifier or a unique
        prefix; an unresolvable or ambiguous reference is rejected before anything is written.
  \item Because a reference must resolve at write time, no event can refer to a later one, so
        chains of references cannot form cycles.
  \item The retired set is computed at read time as the set of referenced identifiers. Retired
        events stay in the log unchanged, are omitted from the summary and from the context and
        precheck outputs an agent reads, and are labelled \texttt{(superseded)} in search results.
  \item A chain of reversals resolves to its latest decision because every intermediate is itself
        retired. The graph of references is not traversed, and a retired decision is withheld rather
        than presented alongside its replacement.
  \item The target's type is not checked, and nothing prevents two decisions from naming the same
        target; if that happens, both remain standing.
\end{itemize}
The relation is only as complete as the actor makes it: a reversal recorded without a reference
leaves two decisions standing (Section~\ref{sec:limits}).

\paragraph{A short example.} Appendix~\ref{app:example} follows one fictional project through this
lifecycle using the tool's unedited output. In brief: a webhook bug is logged as an issue, two
approaches are recorded as failed attempts, and a third is recorded as the fix; later, a decision to
store idempotency keys in Redis is replaced by one that names it in its \texttt{supersedes} field.
The projected summary then lists the issue with both failed attempts and only the second decision,
and when the agent next prepares to edit the webhook handler, the precheck returns the two approaches
that already failed there. Abridged, the log reads:

\par\noindent
\begin{minipage}{\linewidth}
\begin{lstlisting}[style=pm,basicstyle=\ttfamily\footnotesize]
{"type": "issue",    "issue_id": "0001", "summary": "Webhook retries create duplicate charges"}
{"type": "attempt",  "issue_id": "0001", "outcome": "failed", "summary": "Deduplicated on the Stripe event id"}
{"type": "attempt",  "issue_id": "0001", "outcome": "failed", "summary": "5-second lock around mark_paid"}
{"type": "fix",      "issue_id": "0001", "summary": "Write the idempotency key before charging"}
{"type": "decision", "id": "evt_ff7d...", "summary": "Store idempotency keys in Redis"}
{"type": "decision", "supersedes": "evt_ff7d...", "summary": "Enforce idempotency in Postgres"}
\end{lstlisting}
\end{minipage}

\paragraph{Durability.} Every write, whether from the CLI, a git hook, or an MCP tool, goes through
one function that normalizes the timestamp, applies secret redaction (Section~\ref{sec:impl}),
appends one JSON line to \texttt{events.jsonl}, and considers the event for the cross-project store;
the summary is then regenerated. Two properties follow that a reader should not assume away. There
is no inter-process lock, so concurrent writers are not coordinated. And the log and its projection
are updated in separate, non-atomic steps: a crash between them leaves the summary stale until the
next write or an explicit \texttt{pjm regenerate}, while a crash during an append can leave a torn
final line, which the reader rejects with an error naming the line until it is repaired by hand.

% ═══════════════════════════════════════════════════════════
\section{Implementation}
\label{sec:impl}
\label{sec:arch}
\label{sec:capabilities}

\begin{figure}[!t]
\centering
\includegraphics[width=\linewidth]{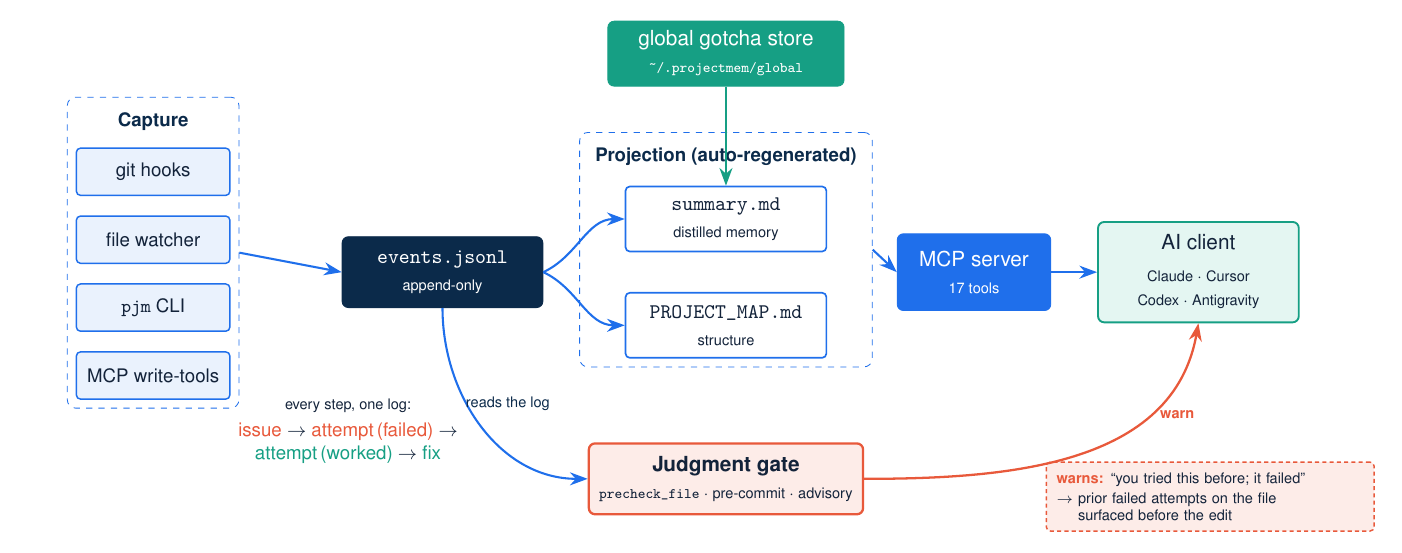}
\caption{projectmem data lifecycle. Four capture sources append to one append-only event log;
deterministic projections distill it into AI-readable files; a native MCP server serves them to
clients; and an advisory judgment gate reads the same log to surface a file's prior failures before
an edit. A machine-wide store carries library lessons across projects.}
\label{fig:arch}
\end{figure}

The implementation described in this section is projectmem v0.3.3.
Figure~\ref{fig:arch} shows the data lifecycle. Four capture sources feed a single event log;
deterministic projections distill it; an MCP server exposes it to AI clients; and the judgment gate
reads the same log to warn before an edit. A machine-wide store carries library-level lessons across
projects. projectmem is implemented in Python ($\geq 3.10$), published on PyPI, and installed with
\texttt{pip install projectmem} followed by \texttt{pjm init}. It has three runtime dependencies
(Typer, the MCP SDK, and watchdog); the package itself occupies about 1.5\,MB installed, excluding
those dependencies. There is no database engine; the
system is files plus a stdio server.

\paragraph{Capture.} Memory is populated with limited manual bookkeeping. Git hooks classify each
new commit by its message in the background: a revert becomes a failed attempt, a fix commit a fix,
a refactor or breaking change a decision, and a feature a note, while messages that match no pattern
are skipped. An opt-in file watcher records rapid, repeated edits to one file. The \texttt{pjm} CLI
(26 top-level commands and one command group, plus one hidden command that the git hooks call) and the MCP write tools let a developer or
an agent record events directly. On initialization, \texttt{pjm init} backfills memory from recent
git history and reads the project's manifests (\texttt{pyproject.toml}, \texttt{package.json},
\texttt{Cargo.toml}, \texttt{go.mod}) to seed the project map, so an existing repository does not
start with empty memory.

\paragraph{MCP server.} The server is built on FastMCP and exposes \textbf{17 tools: 12 read and
5 write}. The read tools cover the session-start summary and instructions, the project map, issue
lookup, event search, scoring, token-budgeted context, cross-project lessons, the plan file, the
project registry, and the \texttt{precheck\_file(file\_path)} gate. The write tools mirror the event
schema: \texttt{log\_issue}, \texttt{record\_attempt}, \texttt{record\_fix}, \texttt{add\_decision},
and \texttt{add\_note}. Every tool body runs under a wrapper that turns exceptions into readable text
rather than crashing the session, and inside a stdout-suppression context so that stray output
cannot corrupt the JSON-RPC stream. Tool parameters carry JSON-schema constraints
(\texttt{search\_events.limit} is bounded to $[1,100]$, \texttt{record\_attempt.outcome} is checked
against \texttt{worked|failed|partial}, and \texttt{get\_context.tokens} is bounded to
$[100, 20000]$), so malformed calls are rejected at the schema layer.

Since v0.3.0 one server serves every registered project, so each call resolves its own project rather
than the server being pinned to a repository at launch. Resolution is a fixed precedence, and the
order encodes a safety property rather than a convenience. A server started with \texttt{-{}-root}
is a boundary, not a default, so it outranks even an explicit argument, and a mismatched argument is
an error rather than a silent redirect: an agent told it wrote to \texttt{beta} must never be left
believing that when the write went elsewhere. Below the boundary come an explicit \texttt{project=}
argument, the active project, the working directory (as git locates \texttt{.git/}), and, when
exactly one project is registered, that project. Failing all of these, the server refuses to guess
and says so, since a wrong guess corrupts an audit trail that is meant to be replayable.\footnote{A
further position is reserved for MCP client roots, which would let a host name the project it has
open. The MCP session exposes them, but the synchronous tool bodies cannot yet await that call, so
the branch is present and never taken; we describe the order as implemented.}

\paragraph{Cross-project operation.} projectmem works across projects in three ways, which differ
in what they share. First, one MCP server serves every registered project, as described above, while
each project's memory stays in its own repository. Second, the global dashboard (\texttt{pjm
dashboard}) reads every registered project's log at the moment it is generated and presents each
project's grade, issues, fixes, and estimated savings side by side. It is a view, not a store: nothing is copied into a
central database, and projects are found through the opt-in registry rather than by searching the file
system. Third, lessons that concern a library rather than a single repository are promoted to a
machine-wide store under \texttt{\textasciitilde/.projectmem/global/} and resurfaced, with the name of
the project they came from, in a later project that uses the same dependency
(Figure~\ref{fig:global}). Promotion is gated by a signal filter: failed or partial attempts always
qualify, because the outcome is the signal, while decisions and notes qualify only when explicitly
marked as lessons (\texttt{gotcha:}, \texttt{lesson:}, \texttt{avoid:}, \texttt{never}, and similar).
A qualifying lesson is attributed to a library when that library's name occurs in its text as a whole
word, drawn from a cache of every package name the machine's manifests have declared. When the
project itself declares dependencies, the match must also be one of them, so a React project's
lessons stay out of a Go project's context; when it declares none, this check does not apply.
Section~\ref{sec:eval} shows the consequence. Nothing in the store is synchronized off the machine.

\begin{figure}[!t]
\centering
\begin{tikzpicture}[
  font=\footnotesize, node distance=5mm,
  b/.style={draw,rounded corners=2pt,align=center,inner sep=4pt,minimum height=8mm,line width=0.6pt},
  pa/.style={b,fill=pmtint,draw=pmblue},
  filt/.style={b,fill=pmred!10,draw=pmred},
  g/.style={b,fill=pmteal,text=white,draw=pmteal},
  pb/.style={b,fill=pmtint,draw=pmblue},
  warn/.style={b,fill=pmred!12,draw=pmred,text=pmnavy},
  a/.style={-{Latex[length=2mm]},draw=pmteal,line width=0.8pt},
]
\node[pa] (pa) {\textbf{Project A}\\\scriptsize failed attempt on \texttt{vite}\\\scriptsize or note ``gotcha: \ldots''};
\node[filt,right=8mm of pa] (filt) {signal filter\\\scriptsize + stack / library match};
\node[g,right=8mm of filt] (glob) {global store\\\scriptsize \texttt{\textasciitilde/.projectmem/global}};
\node[pb,right=8mm of glob] (pb) {\textbf{Project B}\\\scriptsize same stack (\texttt{vite})};
\node[warn,below=7mm of pb] (warn) {\texttt{get\_global\_gotchas}\\\scriptsize warns, w/ \texttt{source\_project}};
\draw[a] (pa) -- (filt);
\draw[a] (filt) -- (glob);
\draw[a] (glob) -- (pb);
\draw[a] (pb) -- (warn);
\draw[a,draw=pmblue] (glob.south) to[out=-90,in=180] (warn.west);
\end{tikzpicture}
\caption{Cross-project memory. A library-level lesson logged in one project is filtered for signal,
promoted to a machine-wide store keyed by stack, and surfaced, with source attribution, in a later
project that uses the same library. The store never leaves the machine.}
\label{fig:global}
\end{figure}

\paragraph{Privacy and secret redaction.} The system is local: there is no telemetry, and no network
request is made unless the user explicitly enables the update check (\texttt{pjm doctor --online}
for a single check, or \texttt{--auto} for at most one a day) or clicks the ``check for updates''
link in a dashboard, which fetches PyPI's public version information from the browser. Because the
log is plain text that is often committed to git, an accidentally pasted credential would otherwise
persist on disk, so every event's user-supplied text is scrubbed before it is written, with matches
replaced by \texttt{[REDACTED:\textless kind\textgreater]} and a notice emitted. Patterns are anchored
to recognizable, minimum-length prefixes (OpenAI and Anthropic \texttt{sk-} keys, GitHub tokens, AWS
\texttt{AKIA} identifiers, Google \texttt{AIza} keys, Slack and Stripe tokens, JWTs,
\texttt{Bearer} tokens, and PEM private-key headers) and are designed not to alter ordinary
debugging prose; true- and false-positive tests pin that behaviour, but pattern matching offers no
guarantee in either direction. The redactor is also fail-open by design: if the scrubber itself
raises an error, the event is written unredacted, trading the risk of a stored secret for the
guarantee that a logging tool never loses an event.

\paragraph{Views for people.} \texttt{pjm brief} prints a one-screen briefing for the start of a
session: active warnings, possibly stale memories, open issues, recent decisions, and lessons for the
project's stack. \texttt{pjm score} summarizes a project's recorded failures, decisions, and fixes,
with the tool's own estimates of hours and tokens saved, in text or JSON. \texttt{get\_context} and
\texttt{pjm wrap} assemble a context block (active warnings, recent decisions, relevant fixes, and the
relevant part of the project map) sized to a token budget, for agents that do not call the MCP tools
directly. \texttt{pjm visualize} renders the log as a local, self-contained HTML dashboard with a
failure heat map by file, the project's structure, and a timeline (Figure~\ref{fig:dashboard} in
Appendix~\ref{app:example}). \texttt{pjm map --build} derives a code-structure graph from the source,
kept outside memory in a disposable cache, over which the dashboard paints each file's failure
history; import relationships are currently resolved only for Python.

\paragraph{Portability.} Git runs hooks under a non-interactive shell that does not read the user's
shell profile, so a bare \texttt{command -v pjm} fails for the many users whose interpreter lives in
a conda, pyenv, or virtual environment. projectmem therefore resolves the absolute path to
\texttt{pjm} at install time and writes it into the hook, with a runtime fallback if the binary has
moved, and the hooks capture in the background so the commit does not wait for capture to finish. A
second issue surfaced on Windows. An unresolvable interpreter line does not cause git to skip a hook;
it causes git to
refuse the commit. The hooks originally declared \texttt{\#!/usr/bin/env bash}, which depends on the
search path and is often unavailable under graphical git clients, and the embedded Windows path lost
its backslashes to shell escaping. The hooks now declare \texttt{\#!/bin/sh}, which Git for Windows
maps to its bundled shell, and embed the path with forward slashes. For MCP clients, \texttt{pjm init}
prints one configuration block, with the absolute interpreter path written in, that serves every
registered project. Because clients differ in which instruction file they read, the repository
bridge is written to both \texttt{CLAUDE.md} and \texttt{AGENTS.md} and names the project, so that
an agent's first tool call can address it directly. For agents that read instructions from the
repository but do not call MCP tools, \texttt{pjm export --claude-md} compiles the standing decisions,
lessons, and failed approaches into \texttt{CLAUDE.md}, so the gate's judgment reaches them in static
form. \texttt{pjm doctor} checks each installed client's configuration and reports servers still pinned
to one repository with \texttt{-{}-root}, configurations it could not read, and a fix that has since
been reverted.

\paragraph{Storage layout and testing.} All per-project state is human-readable text under
\texttt{.projectmem/}, and machine-wide state under \texttt{\textasciitilde/.projectmem/global/}:
\par\noindent
\begin{minipage}{\linewidth}
\begin{lstlisting}
.projectmem/
  events.jsonl        # append-only event log, the source of truth
  summary.md          # deterministic projection the agent reads
  PROJECT_MAP.md      # stack and structure
  AI_INSTRUCTIONS.md  # workflow rules served at session start
  plan.md             # intent: ideas and plans, edited directly
  issues/0042-*.md    # per-issue history (token-efficient reads)
  .current_issue      # marker for issue attribution
~/.projectmem/global/ # cross-project lessons, keyed by library
  .promotable.json    # cache of recognizable libraries
\end{lstlisting}
\end{minipage}
The implementation is covered by \textbf{279 automated tests}. The package declares support for
Python 3.10 and later; we ran the suite on Python 3.11, 3.12, and 3.13, where all tests pass. The
suite includes tests that pin the redactor's behaviour and a test that checks the batched staleness
computation of Section~\ref{sec:eval} against the per-event computation it replaced.

% ═══════════════════════════════════════════════════════════
\section{Evaluation}
\label{sec:eval}

We evaluate projectmem as a system along six empirical facets: estimated token cost, a self-study of
real use, the latency of the judgment gate, the use of supersession in the field, the attribution of
cross-project lessons, and compatibility across clients. We also describe the audit trail the system
produces. We are explicit about what is measured, what is observed, and what is estimated, and about
what the evaluation does not show: we do not measure how often the gate prevents a repeated failure.
Section~\ref{sec:limits} proposes a benchmark for that question.

\paragraph{Estimated token cost.} In MCP mode, an agent's session-start reads (\texttt{get\_summary}
and related calls) amount to roughly 800 to 1{,}500 tokens for a project of moderate size, and the
Markdown bridge to roughly 2{,}500. We estimate that reconstructing the same context with no memory
layer costs 5{,}000 to 20{,}000 tokens per session, consistent with the per-turn long-context costs
analyzed by~\citet{beyondcontextwindow2026}. These are order-of-magnitude estimates from our own use,
not measurements taken with a fixed tokenizer and protocol, and we made no paired comparison, so we
report no percentage reduction. We include these estimates only as motivation and do not use them as
evidence of token savings. The session-start read is also not fixed: it grows with the number of
decisions a project records (Section~\ref{sec:limits}).

\begin{figure}[t]
\centering
\includegraphics[width=0.8\linewidth]{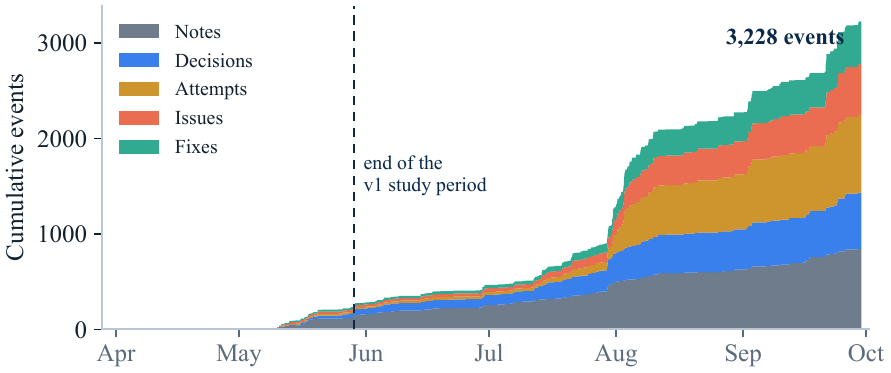}
\caption{Cumulative events by type across 27 projects on two machines (real event logs,
$N=3{,}228$, 30 March to 29 September 2026). The dashed line marks the end of the period reported in
the first version of this paper.}
\label{fig:growth}
\end{figure}

\begin{figure}[t]
\centering
\begin{subfigure}{0.49\linewidth}\centering
  \includegraphics[width=\linewidth]{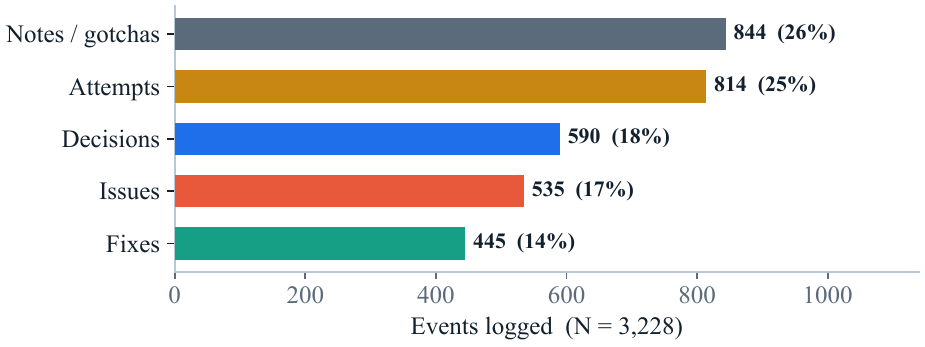}
  \caption{Event-type composition ($N=3{,}228$).}
\end{subfigure}\hfill
\begin{subfigure}{0.49\linewidth}\centering
  \includegraphics[width=\linewidth]{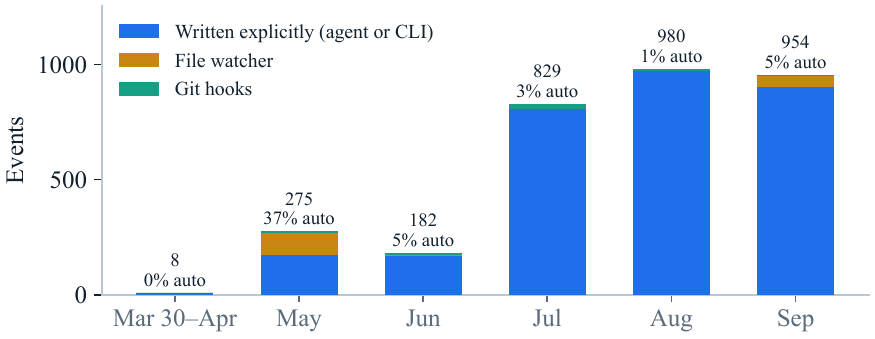}
  \caption{Events by month and how they were captured.}
\end{subfigure}
\caption{Captured memory in the self-study. Issue-lifecycle events---issues, attempts, and
fixes---make up 56\% of events, and 94\% of events were written explicitly rather than captured
automatically.}
\label{fig:dist}
\end{figure}

\begin{figure}[t]
\centering
\includegraphics[width=0.66\linewidth]{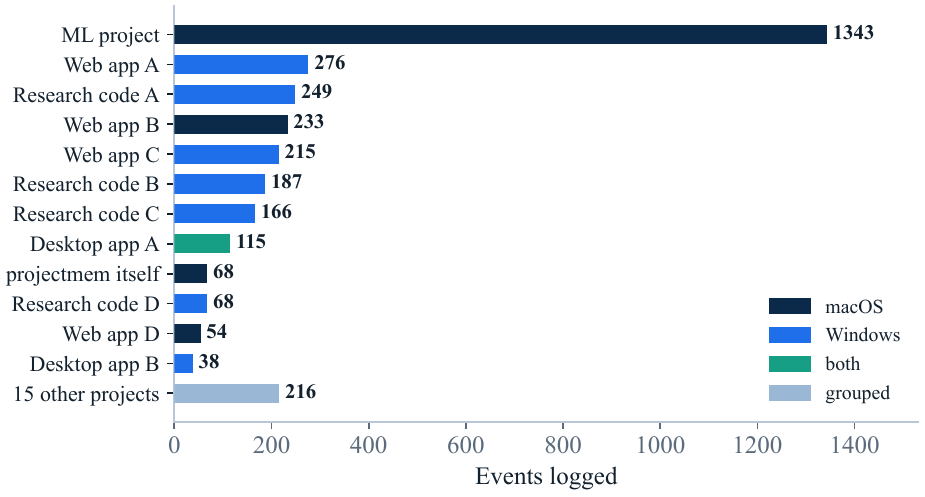}
\caption{Events per project in the self-study, anonymized. The twelve largest projects are shown,
coloured by the machine on which they were worked; category labels were assigned by inspecting each
project, and letters distinguish projects of the same category.}
\label{fig:projects}
\end{figure}

\paragraph{Self-study.} We instrumented our own development on two machines, one running macOS and one
running Windows, over six months (30 March to 29 September 2026). To collect the data we scanned
every folder under the user profile on each machine, and a second data drive on the Windows machine,
for projectmem event logs; excluded test beds, test fixtures, and demonstration projects; kept the
events timestamped within the study period; and counted each event once by its identifier, since
projects synchronized between the machines or copied between folders share events. This yields
\textbf{3{,}228 events from 27 projects}, spanning machine learning, web and desktop applications,
developer tools, and research code. The first version of this
paper reported 207 events from ten projects over the first two months; for the same period, the wider
collection finds 275 events from eleven projects. Figure~\ref{fig:growth} shows how the events
accumulated by type, Figure~\ref{fig:dist} their composition and how they were captured, and
Figure~\ref{fig:projects} their distribution across the anonymized projects.

The system evolved during the study: releases changed the schema, agent instructions, capture
coverage, and read paths. Comparisons between early and late periods are therefore descriptive and
may reflect instrumentation and policy changes as well as changes in use.

Three observations follow. First, issue-lifecycle events---issues, attempts, and fixes---make up 56\%
of events, against 14\% in the period reported in the first
version, with decisions at 18\% and notes at 26\%. Second, capture is overwhelmingly explicit: 94\% of
events were written through the CLI or the MCP write tools, and 6\% were captured automatically, by
the file watcher (145 events) and the git hooks (42). The log does not record who issued a write, so
we cannot separate an agent's writes from a developer's. Third, the issue records show how often the
gate had a prior failure on record to report (Table~\ref{tab:lifecycle}). We counted 529 distinct
\texttt{(project, issue\_id)} pairs; the logs contain six additional duplicate issue events. Of these
distinct issues, 427 (81\%) were closed by a recorded fix, and 86 of those (20\%) had at least one
failed attempt earlier than the first fix in event-log order. Such a failed attempt is eligible for a
later gate warning only when it is attributed to the same file and falls within the gate's lookback
window. Counting partial attempts, which the summary carries but the gate does not warn on, the share
rises to 42\%. These are candidate histories for the gate, not warnings shown or heeded.

The data are concentrated and come from our own work: the largest project contributes 42\% of events
and the ten largest 90\%, while eight projects have fewer than ten events each. The self-study shows
that the log accumulates structured project memory in the authors' routine use, on two operating systems. It does
not show how often the gate fired, whether its warnings were relevant, or whether they changed what an
agent did: we did not instrument those, and we make no causal claim about productivity.

\begin{table}[t]
\centering
\caption{Issue lifecycle in the self-study. Distinct issues are counted by
\texttt{(project, issue\_id)}. An attempt precedes a fix when it appears earlier in event-log order;
log order breaks ties when timestamps have one-second resolution. A failed attempt can be reported
only on a later edit attributed to the same file and within the gate's lookback window.}
\label{tab:lifecycle}
\small
\begin{tabular}{lr}
\toprule
Distinct issues opened & 529 \\
\quad closed by a recorded fix & 427 (81\%) \\
\quad\quad with a failed attempt before the fix & 86 (20\% of fixed) \\
\quad\quad with a failed or partial attempt before the fix & 180 (42\% of fixed) \\
Failed attempts before the fix, per such issue (median; maximum) & 1; 10 \\
Attempts by outcome (worked; partial; failed) & 418; 227; 169 \\
\bottomrule
\end{tabular}
\end{table}

\paragraph{Gate latency (measured).} Because the gate is meant to be called before every edit, its
cost is paid continually, and it should not grow with the age of a project. In v0.3.2 it did. To
decide whether a memory citing a file had gone stale, the gate issued one \texttt{git log} per event,
computed staleness over the entire log, and only then filtered the result to the file being checked.
v0.3.3 restricts the computation to the files under check and issues one \texttt{git log} per
distinct file, bounded to commits newer than the oldest event that cites it, with the remaining
comparison done in memory. Table~\ref{tab:latency} reports both versions on the same synthetic
repository (four source files and 60 commits spanning 200 days) with logs of 100, 400, and 1{,}500
events. Each configuration was run five times after one warm-up run, and we report the median and
range of the analysis time for one file, excluding interpreter start-up. The earlier gate's time and
subprocess count grow in proportion to the log, reaching a median of 42.0\,s and 1{,}500
\texttt{git} invocations at 1{,}500 events. The revised gate issues two invocations at every size and
completes in under 83\,ms; its cost is nearly constant within the measured range, and the remaining
in-memory work, which still grows with the log, accounts for the rise from 57 to 79\,ms. Both
versions raise the same failed-attempt warning in every configuration, and an automated test checks
that the batched staleness count equals the per-event computation, including across a merge commit.
All measurements were taken on one workstation (Intel Core i5-7500, four cores, macOS 13.7,
Python 3.12.2, git 2.39.3). Absolute times will differ elsewhere. The fixed count of two git
subprocesses follows from the revised algorithm, while its remaining in-memory scan still grows with
the event log.

\begin{table}[t]
\centering
\caption{Judgment-gate latency for one file against event-log size, on an identical synthetic
repository: median and range over five runs of the analysis time, excluding interpreter start-up, and
the number of \texttt{git} subprocesses per check.}
\label{tab:latency}
\small
\begin{tabular}{rrrrr}
\toprule
 & \multicolumn{2}{c}{v0.3.2} & \multicolumn{2}{c}{v0.3.3} \\
\cmidrule(lr){2-3}\cmidrule(lr){4-5}
Events & Median (range), ms & \texttt{git} calls & Median (range), ms & \texttt{git} calls \\
\midrule
100     & 2{,}748 (2{,}736--2{,}827)    & 100     & 56.7 (56.2--60.5) & 2 \\
400     & 11{,}286 (11{,}167--11{,}533) & 400     & 61.8 (59.8--65.6) & 2 \\
1{,}500 & 42{,}047 (41{,}007--42{,}296) & 1{,}500 & 79.3 (74.9--82.6) & 2 \\
\bottomrule
\end{tabular}
\end{table}

\paragraph{Supersession in field use (observed).} The supersession reference was released in
v0.1.4, shortly after the first version of this paper was submitted, and was fully specified in the
schema of the \texttt{add\_decision} tool: the parameter's description stated what it did and when to
use it. It was not, however, mentioned on any of the three surfaces an agent reads at the start of a
session: the instruction file (\texttt{AI\_INSTRUCTIONS.md}, 12{,}503 characters), the repository
bridge file, or the MCP server's own instructions. The schema also tied its use to a different
trigger, a staleness warning from the gate, rather than to a change of mind. Table~\ref{tab:supersession}
counts supersession references in the self-study logs by period. While the field existed but was
undocumented, 9 of 390 decisions carried one, and 8 of those 9 came from a single project within six
hours, in one session in which an agent found the field in the schema and used it repeatedly. We do
not know how many decisions reversed an earlier one, so these are counts rather than usage rates; they
show only that the relation was rarely written. The practical consequence was reported by
an external user: summaries accumulated decisions that contradicted one another, with nothing to
indicate which was current. v0.3.3 described the relation on all three surfaces, stated what it does
to the summary rather than only that the parameter exists, and attached it to the moment of reversal.
We then ran a small scripted reversal task once in each of two MCP clients: an agent was asked to
build a small web interface in one language and then told to switch to another. In both runs the
agent recorded the second decision with a reference to the first without being asked to, and in one
it first searched the event log for the identifier of the decision it was about to retire. Two runs
are a demonstration rather than a measurement, and the field logs do not yet corroborate them: after
the change, 4 of 119 decisions carried a reference, but 17 days of data are too few to show an effect
in either direction. The field data establish only that the relation was rarely written while it
appeared solely in the schema, and the scripted runs that agents can write it unprompted once it is
described. Together they are consistent with a practical lesson for agent-facing tool design: a
capability that is correctly specified in a tool's interface, but absent from the instructions an
agent reads, may go largely unused.

\begin{table}[t]
\centering
\caption{Decisions and supersession references in the self-study, by period. The field was added in
v0.1.4 (11 June 2026) and first described in the agent's instructions on 13 September 2026. The
boundaries are those of the change on the development machine; the Windows machine received each
release some days later.}
\label{tab:supersession}
\small
\begin{tabular}{lrr}
\toprule
Period & Decisions & With \texttt{supersedes} \\
\midrule
30 March to 11 June (field not yet available) & 81 & 0 \\
12 June to 12 September (available, undocumented) & 390 & 9 \\
13 to 29 September (described in the instructions) & 119 & 4 \\
\bottomrule
\end{tabular}
\end{table}

\paragraph{Cross-project attribution audit.} At the end of the study the machine-wide store on the
macOS machine held 46 lessons from six projects, attributed to 15 libraries. One author read each
lesson and judged whether it concerned the library it was filed under. Twelve (26\%) did, including
lessons about PyTorch, pandas, Prisma, and Stripe; five (11\%) were ambiguous, four of them naming
Python as a language rather than a library; and 29 (63\%) did not. Of those 29, 28 had been attributed to a package whose name is
also an ordinary English word (\texttt{full}, \texttt{match}, \texttt{build}, \texttt{next},
\texttt{globals}, and \texttt{raster}), matched in phrases such as ``did not match'' and ``a clean
build''. All 28 came from the two projects that declare no dependencies in a manifest, where the check
against the project's own dependencies does not apply (Section~\ref{sec:impl}); the projects that do
declare them contributed no such lesson. Whole-word matching against package names therefore
misattributes lessons whenever a package shares its name with a common word and nothing else
constrains the match, so a lesson about a document build can surface in any project that depends on a
package called \texttt{build}. This is a defect in the implementation, and one author's classification
of 46 lessons from one machine is a small sample; we report it because it bounds how far the
cross-project mechanism, as currently built, can be trusted.

\paragraph{Compatibility validation.} For the first version of this paper, we verified the
configuration end-to-end against a real project in four MCP clients. For the present evaluation of
projectmem v0.3.3, we ran a four-step acceptance suite on macOS
(Table~\ref{tab:clients}): starting a session and reaching the right project without a lookup call
(T1); retrieving context for that project (T2); reversing a decision and recording the supersession
(T3); and running the gate on a 1{,}500-event log, checking both its cost and its warnings (T4). The
suite was run in three clients; Cursor, included in the first paper version, was not re-run. The
command-line parts of the suite, including installing the git hooks and committing through them, were
repeated on Windows 11 with Python 3.12.10 and Git 2.54.0.

\begin{table}[t]
\centering
\caption{Acceptance results for projectmem v0.3.3.}
\label{tab:clients}
\small
\begin{tabular}{llllll}
\toprule
Client & Version & T1 & T2 & T3 & T4 \\
\midrule
Antigravity                & 2.1.4        & pass    & pass    & pass           & pass \\
Claude desktop app (Code tab) & 2.9939.2 & resolution only & pass & read path only & pass after fix \\
Codex                      & 26.908.40834 & pass    & pass    & pass           & pass \\
Cursor                     & 3.4.17       & not run & not run & not run        & not run \\
\bottomrule
\end{tabular}
\end{table}

The macOS runs took place from 13 to 15 September 2026; Table~\ref{tab:clients} reports the client
versions installed for each run, although clients may have updated between runs. The Claude desktop
app was tested through its Code tab. ``Resolution only'' means that a deliberately issued call
without a project argument resolved to the right project, but we did not observe an agent choose the
call unprompted. ``Read path only'' means that the client displayed the projected result of a
reversal recorded elsewhere but did not author the reversal. ``Pass after fix'' means that the run
exposed the two defects described below and passed after they were corrected.

Cross-client testing proved necessary for correctness, not only for compatibility. Clients differ in
the working directory from which they launch the server: some start it inside the project, others
elsewhere. Two defects were invisible both to clients of the first kind and to the command line, where
the two directories coincide: the context tool read repository state from the server's own
directory, and the gate judged staleness and churn against it. Both surfaced only in a client of the
second kind. We therefore recommend that memory servers be tested under both launch conventions.

\paragraph{Auditability (descriptive).} Because captured events are timestamped plain text, the log doubles as a
provenance trail. This is relevant to the reproducibility concerns raised for LLM-assisted software
engineering~\citep{llm4se_reproducibility2025}: a record of what was changed, what was tried, and why
it was done accumulates as a byproduct of normal work and is reviewable with standard version-control
tools. The trail is only as complete as capture, however: commits whose messages match no pattern are
skipped, and an agent can omit a write it was instructed to make.

% ═══════════════════════════════════════════════════════════
\section{Related Work}
\label{sec:related}

We organize prior work into four threads (retrieval-oriented agent memory, project memory for
coding agents, learning from failure, and pre-action guardrails) and then state the gap that
projectmem fills.

\paragraph{Retrieval-oriented agent memory.} The dominant paradigm extracts, consolidates, and
retrieves salient context into vector and/or graph stores to overcome a fixed context window.
\citet{li2026memcog} characterize essentially all such systems as Memory-as-Tool: one query
in, one flat top-$k$ list of passages out. MemGPT/Letta pages context in and out like an operating
system over a tiered, mutable memory with function-call self-editing~\citep{packer2023memgpt};
Mem0 dynamically extracts and vector-indexes salient conversational facts, its graph variant Mem0g
adding only marginal accuracy~\citep{chhikara2025mem0}; A-MEM organizes memory as a Zettelkasten
knowledge network of LLM-generated, dynamically-linked notes~\citep{xu2025amem}; Zep/Graphiti is a
temporally-aware knowledge-graph engine retrieved via embeddings and reranking~\citep{rasmussen2025zep};
and MemMachine combines a vector database and a graph database (and exposes an MCP server), explicitly
optimizing retrieval accuracy~\citep{wang2026memmachine}. Most recently, Memanto pairs a
typed semantic-memory schema with information-theoretic retrieval, reporting state-of-the-art
accuracy on the LongMemEval and LoCoMo QA suites via single-query reads~\citep{abtahi2026memanto}.
Its client is open-source, but retrieval is delegated to a hosted service (Moorcheh), so Memanto is
neither local nor offline. We view it as independent evidence for
typed memory, aimed at retrieval fidelity rather than the local-first, plain-text, action-gating
design we pursue. Generative Agents introduced the
append-only natural-language memory stream with recency/importance/relevance retrieval and
reflection~\citep{park2023generative}. It is an intellectual ancestor of our event log, though its
retrieval is LLM-scored and lossy where ours is deterministic and exact. MemoryBank, by contrast,
deliberately forgets via an Ebbinghaus curve~\citep{zhong2024memorybank}, the opposite of our
never-forget audit trail. The cognitive-architecture framing of CoALA~\citep{sumers2023coala} maps
classical working/episodic/semantic/procedural memory onto LLM agents. These systems are
predominantly embedding- or graph-backed, conversational or personalization-oriented, frequently
cloud-hosted, typically mutable, and evaluated on QA benchmarks; crucially, the reviewed systems primarily augment
context rather than being consulted at the point of an action. A recent benchmark of memory in LLM agents finds that no
current architecture masters all of accurate retrieval, test-time learning, long-range
understanding, and conflict resolution~\citep{hu2025memoryagentbench}, which is independent evidence that the
design space remains unsaturated.

\paragraph{Stale and superseded memory.} A growing line of work treats the validity of a memory, and
not only its retrieval, as the problem. STALE shows that frontier models and memory frameworks commonly
fail to notice when a later observation implicitly invalidates an earlier memory~\citep{chao2026stale}.
Where systems handle outdated memory, they usually do so by inference: Mem0 asks a model to decide
whether a new fact adds to, updates, or deletes an existing one~\citep{chhikara2025mem0}, and Zep
invalidates a knowledge-graph edge when a model judges that newer information contradicts
it~\citep{rasmussen2025zep}. MemStrata instead retires a stale value deterministically, by a
supersession rule keyed on an extracted (subject, relation, object) triple~\citep{yadav2026memstrata};
on 130 state transitions taken from real GitHub fixes, retrieval-augmented generation served the
superseded value in 36 to 38\% of forced answers, against near zero for
MemStrata~\citep{yadav2026softwarehistories}. TOKI formalizes contradiction resolution as write-time
concurrency control and preserves the losing fact in an audit row~\citep{wang2026toki}, and MemTX
stages writes in transactions and repairs derived records when a belief is
retracted~\citep{li2026memtx}. Finally, Shen et al.\ find that none of the five systems they study
enforces a revocation mark at retrieval by default, so a revoked fact is still returned and acted
upon~\citep{shen2026revoked}. projectmem had the same fault until v0.3.3, when its context and
precheck outputs began to filter retired decisions. It differs from these systems in where the
relation comes from: the reference that retires one decision is written explicitly by the actor at the
moment of writing, and nothing infers it from content, at write time or at read. We discuss the cost
of that choice in Section~\ref{sec:limits}.

\paragraph{Project memory for coding agents.} A smaller, very recent thread targets persistent memory
for software-engineering agents specifically. Closest to our work is Codified Context, which, like
projectmem and independently, adopts a plain-text, file-based, no-vector-database design served over
an MCP server, motivated by the identical observation that agents ``lose coherence across sessions,
forget project conventions, and repeat known mistakes''~\citep{vasilopoulos2026codified}. It pairs a
Markdown ``hot-memory constitution'' of conventions with on-demand specification documents retrieved
by keyword. The decisive difference is mechanism: Codified Context supplies project knowledge
passively: it provides conventions as context the agent must read, and its drift detector
fires at session start from git-commit divergence, not from the project's logged failures and
not per action. projectmem instead derives a deterministic, per-file warning from an
append-only event log of typed failures. The event-sourcing substrate itself has been
proposed concurrently for autonomous SE agents (an append-only log of intentions and effects from
which state is deterministically projected~\citep{santos2026esaa}), which we read as validation of
the substrate; projectmem adds the judgment layer, cross-project memory, and human-legibility on top.
We note that MCP support is increasingly common (both MemMachine and Codified Context expose MCP
servers~\citep{wang2026memmachine,vasilopoulos2026codified}), so projectmem's contribution rests not
on the protocol but on the combination of local-first, event-sourced plain text, and a
deterministic judgment gate. MCP itself~\citep{anthropic2024mcp} is the standard that makes the
system tool-agnostic. MCP security literature concerns untrusted, network-reachable
servers~\citep{hou2025mcp}; projectmem reduces its exposure to that threat class by being local,
read-mostly, and auditable, although a local server is not free of risk.

\paragraph{Learning from failure (post-hoc).} A line of work has agents learn from their own
mistakes. Reflexion stores verbal feedback on a failed trial in an episodic buffer that improves the
next trial without weight updates~\citep{shinn2023reflexion}, and recent work shows agents can
attribute a cascading failure to its root cause and iteratively recover from
it~\citep{zhu2025wherefail}. These are retrospective: they react after a failure within a task
or across trials. projectmem differs by externalizing related records across sessions and projects,
structuring them into a fixed typed schema, and converting a post-hoc next-trial hint into an
advisory check made before the next edit.

\paragraph{Pre-action guardrails.} A distinct lineage does intercept actions before execution.
AGrail is the closest category match: a boolean pre-action gate that blocks an agent's action, with a
memory module that iteratively optimizes its safety checks over a lifetime~\citep{luo2025agrail}.
ToolSafe performs proactive step-level pre-execution reasoning~\citep{toolsafe2026}, and LlamaFirewall
layers jailbreak, alignment, and insecure-code checks at runtime~\citep{llamafirewall2025};
Meta-Policy Reflexion consolidates reflections into predicate-like rules with admissibility checks at
inference~\citep{metapolicyreflexion2025}. These share projectmem's timing, acting before
the action, but differ in mechanism and objective: their gates are LLM- or RL-trained and
therefore non-deterministic, they typically block rather than advise, and they target generic safety or code-security categories
(jailbreak, prompt injection, insecure code) or administrator-specified risks, not a particular
project's own recorded failed fixes. projectmem's gate is an advisory, deterministic lookup keyed
to the project's failure history; it needs no model call and no training, and it is reproducible.

\paragraph{A third axis: Memory-as-Governance.} Extending the vocabulary of \citet{li2026memcog}, we
distinguish Memory-as-Tool (passive, query-in/passages-out), Memory-as-Cognition
(access interleaved with reasoning), and the cell projectmem occupies, Memory-as-Governance:
memory that is consulted at the point of action and returns a deterministic, advisory warning
derived from the project's history, rather than being read only when the agent chooses to. The
reviewed retrieval systems primarily augment context rather than being consulted at the point of
action; the guardrail
thread acts at that point but on generic, model-trained safety rather than project history; the
coding-agent thread shares our substrate but supplies project knowledge only passively. Among the
systems we reviewed, none is simultaneously (i) local-first and offline, (ii) event-sourced over
append-only, human-readable plain text, (iii) MCP-native, (iv) equipped with a deterministic,
advisory pre-action check derived from the project's own failure history, and (v) cross-project. Table~\ref{tab:compare} situates
projectmem against selected representative systems from each thread.

\begin{table}[t]
\centering
\caption{Capability comparison along properties relevant to AI-assisted software development. This
compares design capabilities, not head-to-head task accuracy; these systems are not measured
on a common benchmark. Entries follow the cited papers; where a paper is silent on deployment or protocol
support, the local-first and MCP columns follow the system's public repository as of September 2026.
Plain text is marked partial for systems that store natural-language records but retrieve them by
embedding similarity. ``Pre-action judgment'' denotes a check made at the point of an action: projectmem's is
advisory, deterministic, and derived from the project's own failure history, distinct from the
guardrails (AGrail, LlamaFirewall), whose gates are model-trained and target generic safety/security
categories. \cmark~yes \quad \pmark~partial \quad \xmark~no \quad n/a~not applicable.}
\label{tab:compare}
\small
\setlength{\tabcolsep}{4.5pt}
\begin{tabular}{lccccccc}
\toprule
\textbf{System} & \makecell{Local-\\first} & \makecell{Plain-text\\(no vector DB)} &
\makecell{Event-sourced\\/ append-only} & \makecell{Pre-action\\judgment} & \makecell{MCP-\\native} &
\makecell{Cross-\\project} & Domain \\
\midrule
MemGPT/Letta~\citep{packer2023memgpt}        & \pmark & \xmark & \xmark & \xmark & \xmark & \xmark & general/chat \\
Mem0~\citep{chhikara2025mem0}                 & \pmark & \xmark & \xmark & \xmark & \xmark & \pmark & chat/personal \\
Gen.\ Agents~\citep{park2023generative}       & n/a    & \pmark & \pmark & \xmark & \xmark & \xmark & simulation \\
A-MEM~\citep{xu2025amem}                      & \pmark & \pmark & \xmark & \xmark & \xmark & \xmark & general \\
Zep~\citep{rasmussen2025zep}                  & \pmark & \xmark & \pmark & \xmark & \pmark & \pmark & chat \\
MemMachine~\citep{wang2026memmachine}         & \pmark & \xmark & \xmark & \xmark & \cmark & \pmark & personal/chat \\
Memanto~\citep{abtahi2026memanto}             & \xmark & \xmark & \pmark & \xmark & \xmark & \xmark & general/QA \\
Reflexion~\citep{shinn2023reflexion}          & \cmark & \cmark & \pmark & \pmark & \xmark & \xmark & tasks/code \\
Meta-Policy R.~\citep{metapolicyreflexion2025}& \cmark & \cmark & \pmark & \cmark & \xmark & \xmark & tasks \\
LlamaFirewall~\citep{llamafirewall2025}       & \cmark & n/a    & \xmark & \cmark & \xmark & \xmark & code security \\
AGrail~\citep{luo2025agrail}                  & \pmark & \xmark & \xmark & \cmark & \xmark & \xmark & agent safety \\
ESAA~\citep{santos2026esaa}                   & \cmark & \cmark & \cmark & \xmark & \xmark & \xmark & SE (pattern) \\
Codified Ctx.~\citep{vasilopoulos2026codified}& \cmark & \cmark & \xmark & \xmark & \cmark & \xmark & AI coding \\
\textbf{projectmem}                           & \cmark & \cmark & \cmark & \cmark & \cmark & \cmark & AI coding \\
\bottomrule
\end{tabular}
\end{table}

% ═══════════════════════════════════════════════════════════
\section{Limitations and Future Work}
\label{sec:limits}

\paragraph{Limitations.} The following limitations bound the claims made above.
\begin{itemize}[leftmargin=1.4em,itemsep=2pt]
  \item \textbf{No measure of prevention.} The evaluation shows that memory accumulates and that the
        gate is cheap in the measured synthetic workloads; it does not show that the gate prevents repeated failures, which requires the
        benchmark proposed below.
  \item \textbf{A file-scoped, advisory gate.} The gate reports which approaches have failed on a
        file; it cannot tell whether a new change repeats one of them. A past failure may no longer
        bear on a superficially similar but valid change, which is why the warning is advisory.
  \item \textbf{History and capture completeness.} The gate is only as good as the logged history;
        on a new project it has little to report, which backfilling from git history mitigates but
        does not remove. A reversal recorded without a supersession reference leaves two decisions
        standing, because projectmem does not infer the missing relation. The hand-edited project map
        can drift from the implementation, the git hooks and watcher rely on unvalidated heuristics,
        and library-name matching can misattribute cross-project lessons (Section~\ref{sec:eval}).
  \item \textbf{No semantic recall.} projectmem deliberately performs no vector retrieval, trading
        fuzzy recall for determinism and zero read-time model cost; it complements rather than
        replaces embedding-based memory.
  \item \textbf{Summary growth.} Notes are capped in the projection, but issues and standing
        decisions are not, so the summary grows with a project's history. In one of our own
        long-lived projects, with 101 standing decisions, it reached roughly 120{,}000 characters,
        on the order of 30{,}000 tokens, well above the session-start estimate of
        Section~\ref{sec:eval}.
  \item \textbf{Durability and redaction.} Writes are not coordinated across processes, the log and
        its projection are updated non-atomically, and a torn final line must be repaired by hand
        (Section~\ref{sec:design}). Secret redaction is pattern-based and fail-open, so it can miss a
        secret in an unfamiliar format and, if the scrubber fails, lets an event through unredacted.
  \item \textbf{Evidence and policy validity.} The schema, agent instructions, capture coverage, and
        read paths changed during the self-study, so temporal differences are descriptive rather
        than causal. The gate's thresholds are engineering defaults whose utility has not been
        evaluated. The self-study covers our own projects and is dominated by one project, which
        contributes 42\% of events; the latency result comes from a synthetic repository on one
        machine; and the client suite is small and partly single-run. Token figures are estimates,
        Table~\ref{tab:compare} compares design properties rather than task accuracy, and the system
        is single-user and local.
\end{itemize}

\needspace{6\baselineskip}
\paragraph{Future work.} Several directions matter most.
\begin{itemize}[leftmargin=1.4em,itemsep=2pt]
  \item \textbf{A controlled repeat-failure benchmark.} The most important missing result is a
        measured prevention rate. Seeding projects with known failed approaches and running agents,
        with and without the gate, on tasks that invite repeating them would measure the fraction of
        repeats avoided, the rate of irrelevant warnings, and the added latency. To our knowledge the
        dev-tool memory category currently lacks such a benchmark.
  \item \textbf{Instrumenting the gate.} An opt-in, local record of each check and the edit that
        followed would show whether warnings are heeded, and is a prerequisite for the benchmark
        above.
  \item \textbf{Complete supersession.} Resolving chains of references by traversal, presenting a
        retired decision alongside its replacement rather than withholding it, detecting two
        decisions that retire the same target, and a token-budgeted projection that keeps the
        session-start read fixed as a project ages.
  \item \textbf{Reliable cross-project attribution.} Declining to promote lessons in projects that
        declare no dependencies, and excluding package names that are common words, would address the
        misattribution measured in Section~\ref{sec:eval}.
  \item \textbf{Judgment at the tool-call boundary.} The gate currently keys on the file and is
        reached when an agent calls it or at commit time; moving it to the agent's tool call, and
        reasoning over the specific change, would let it warn when an edit begins to resemble a
        failed one.
  \item \textbf{An open event schema.} Publishing the event format as a versioned specification
        would let other tools read and write the same log, and would support migration from other
        memory systems in both directions.
  \item \textbf{Shared project memory.} A conflict-free merge of append-only logs, in the spirit of
        local-first software~\citep{kleppmann2019localfirst}, would let a team share one project
        memory over its existing git remote, with no central server.
\end{itemize}

% ═══════════════════════════════════════════════════════════
\section{Conclusion}
\label{sec:conclusion}

Persistent project context remains a practical challenge in AI-assisted software development:
decisions, failed attempts, and rationale held only in one session may not be available to the next.
This paper presented projectmem, a local-first memory and judgment layer that combines an append-only
log of typed development events, deterministic plain-text projections, an MCP interface shared across
clients, and a file-scoped advisory gate consulted before an edit. Its central design contribution,
Memory-as-Governance, uses recorded project history not only for retrieval but also to produce
deterministic warnings at the point of action.

The evaluation establishes feasibility rather than prevention effectiveness. During a six-month
self-study, projectmem accumulated 3{,}228 events across 27 projects on two machines. On a synthetic
repository, batching git queries reduced the median time for a 1{,}500-event precheck from 42.0\,s to
79\,ms while preserving the warnings returned. Field observations and two scripted reversal runs
also showed that a schema capability may remain rarely used until it is described in the instructions
an agent reads. The study did not measure whether the gate prevents repeated failures, and its field
evidence comes primarily from the authors' own projects; a controlled repeat-failure benchmark is
therefore the next evaluation step. More broadly, the results suggest that pre-action repository
queries should scale with the files inspected rather than the number of recorded events, and that
agent-facing capabilities must be exposed on the instruction surfaces agents actually consult.

\section*{Acknowledgments}
We thank Mike Hanley, Vivaan Dhawan, and GitHub contributor @medium-effort for code contributions and
for defect reports from real machines, several of which are described in Section~\ref{sec:eval}. We
thank Sefat E. Rahman for an architectural suggestion concerning the project map. We also thank the
University of Utah, and the open-source communities behind the Model Context Protocol, Typer,
watchdog, and D3.js. projectmem was built independently as open-source software-engineering
infrastructure.

\section*{Availability}
Source code: \url{https://github.com/riponcm/projectmem} (MIT-licensed). Package:
\texttt{pip install projectmem}. Documentation: \url{https://projectmem.dev}.

\paragraph{Reproducibility artifacts.} We will release the benchmark code, raw latency results,
supersession-counting script, anonymized self-study records, and acceptance checklist under
\texttt{paper/v2/}. The released data retain the metadata needed to reproduce our analyses but omit
event text, project names, and file paths to protect private projects.

\appendix
\section{Client Configuration Summary}
\label{app:setup}

projectmem is configured once, not once per repository. After
\texttt{pip install projectmem}, running \texttt{pjm init} in a repository initializes the local
memory directory, registers the project, and prints a single MCP server block to paste into the
client. Every repository initialized afterwards is reachable through that same entry, with no second
server and no client restart. (Before v0.3.0 each repository needed its own pinned server entry;
configurations still carrying \texttt{-{}-root} are reported by \texttt{pjm doctor}.) The canonical server invocation is the Python module over stdio:
\begin{center}
\texttt{python -m projectmem.mcp\_server}
\end{center}
The initializer uses \texttt{sys.executable} to record the absolute interpreter path, which avoids
PATH inheritance problems in hosts that launch MCP servers from non-interactive shells. Each call resolves its own project by the precedence given in
Section~\ref{sec:impl}; a server started with \texttt{-{}-root} remains pinned to one repository and
refuses to write outside it. Client-specific configuration files
change over time, so the public documentation provides the current JSON/TOML blocks and verification
commands: \url{https://projectmem.dev/guide}.

% ═══════════════════════════════════════════════════════════
\section{A Worked Example}
\label{app:example}

This appendix follows one small, fictional project, a payments service called \texttt{checkout-api}, through
the lifecycle described in Sections~\ref{sec:design} and~\ref{sec:impl}. Every listing is the unedited
output of projectmem v0.3.3 run on that project: the event log is shown one line per event exactly as
stored, and we have omitted only summary sections that do not bear on the example.

\paragraph{The event log.} A webhook bug is opened, two approaches fail, and a third fixes it. Later a
design decision is made and then reversed; the second decision names the first in its
\texttt{supersedes} field. The log records both decisions and rewrites neither.

\begin{lstlisting}[style=pm,basicstyle=\ttfamily\scriptsize,breakatwhitespace=true]
{"git_commit": "648c87c", "id": "evt_51ad79e762114458bfcf", "issue_id": "0001", "location": "src/payments/webhook.py:2", "summary": "Webhook retries create duplicate charges", "timestamp": "2026-09-26T04:25:35Z", "type": "issue"}
{"git_commit": "648c87c", "id": "evt_1b494cd4368c4de0adec", "issue_id": "0001", "location": "src/payments/webhook.py:2", "outcome": "failed", "summary": "Deduplicated on the Stripe event id; ids repeat across retries", "timestamp": "2026-09-26T04:25:35Z", "type": "attempt"}
{"git_commit": "648c87c", "id": "evt_71f859469ab547458065", "issue_id": "0001", "location": "src/payments/webhook.py:3", "outcome": "failed", "summary": "5-second lock around mark_paid; narrowed the race, did not close it", "timestamp": "2026-09-26T04:25:35Z", "type": "attempt"}
{"git_commit": "648c87c", "id": "evt_6458f86cc0d040d79111", "issue_id": "0001", "location": "src/payments/webhook.py:2", "summary": "Write the idempotency key before charging", "timestamp": "2026-09-26T04:25:35Z", "type": "fix"}
{"git_commit": "648c87c", "id": "evt_ff7d8add7bf1487dad01", "location": "src/payments/webhook.py", "summary": "Store idempotency keys in Redis with a 24-hour TTL", "timestamp": "2026-09-26T04:25:35Z", "type": "decision"}
{"git_commit": "648c87c", "id": "evt_43fdb8861a8d4bec9448", "location": "src/payments/webhook.py", "summary": "Enforce idempotency with a unique key in Postgres; drop the Redis store", "supersedes": "evt_ff7d8add7bf1487dad01", "timestamp": "2026-09-26T04:25:42Z", "type": "decision"}
\end{lstlisting}

\paragraph{The projected summary.} \texttt{summary.md} is regenerated from the log after every event.
The failed attempts are carried forward under the issue they belong to, and only the standing decision
appears; the Redis decision it retired is absent.

\begin{lstlisting}[style=pm]
## Recent issues
- [DONE] #0001 Webhook retries create duplicate charges [src/payments/webhook.py:2] -> Write the idempotency key before charging [src/payments/webhook.py:2] (fixed)
  - Failed attempt: Deduplicated on the Stripe event id; ids repeat across retries [src/payments/webhook.py:2]
  - Failed attempt: 5-second lock around mark_paid; narrowed the race, did not close it [src/payments/webhook.py:3]

## Decisions
- Enforce idempotency with a unique key in Postgres; drop the Redis store [src/payments/webhook.py]
\end{lstlisting}

\paragraph{The judgment gate.} When the agent is about to modify \texttt{webhook.py}, or
later when the change is committed, the gate returns the approaches that have already failed there.
The warning surfaces the failed approaches so that a new session can avoid retrying the event-id
deduplication a previous one ruled out.\footnote{Shown in the plain-ASCII form projectmem emits on terminals
that cannot display box-drawing characters; on other terminals the rules are drawn with line
characters, and the content is identical.}

\par\noindent
\begin{minipage}{\linewidth}
\begin{lstlisting}[style=pm]
projectmem: Pre-Commit Check
------------------------------------------------------------

  src/payments/webhook.py
    WARN  What already failed here (2 attempts):
           x 5-second lock around mark_paid; narrowed the race, did not close it (just now)
           x Deduplicated on the Stripe event id; ids repeat across retries (just now)

------------------------------------------------------------
1 warning(s). Review before committing.
\end{lstlisting}
\end{minipage}

\needspace{9\baselineskip}
\paragraph{Search.} The retired decision is not deleted. Search still returns it, labelled, so the
history of the choice remains reviewable alongside the choice itself.

\begin{lstlisting}[style=pm]
2026-09-26T04:25:35Z [evt_6458f86cc0d040d79111] fix #0001: Write the idempotency key before charging
2026-09-26T04:25:35Z [evt_ff7d8add7bf1487dad01] decision: Store idempotency keys in Redis with a 24-hour TTL (superseded)
2026-09-26T04:25:42Z [evt_43fdb8861a8d4bec9448] decision: Enforce idempotency with a unique key in Postgres; drop the Redis store
\end{lstlisting}

\begin{figure}[!htbp]
\centering
\includegraphics[width=0.8\linewidth]{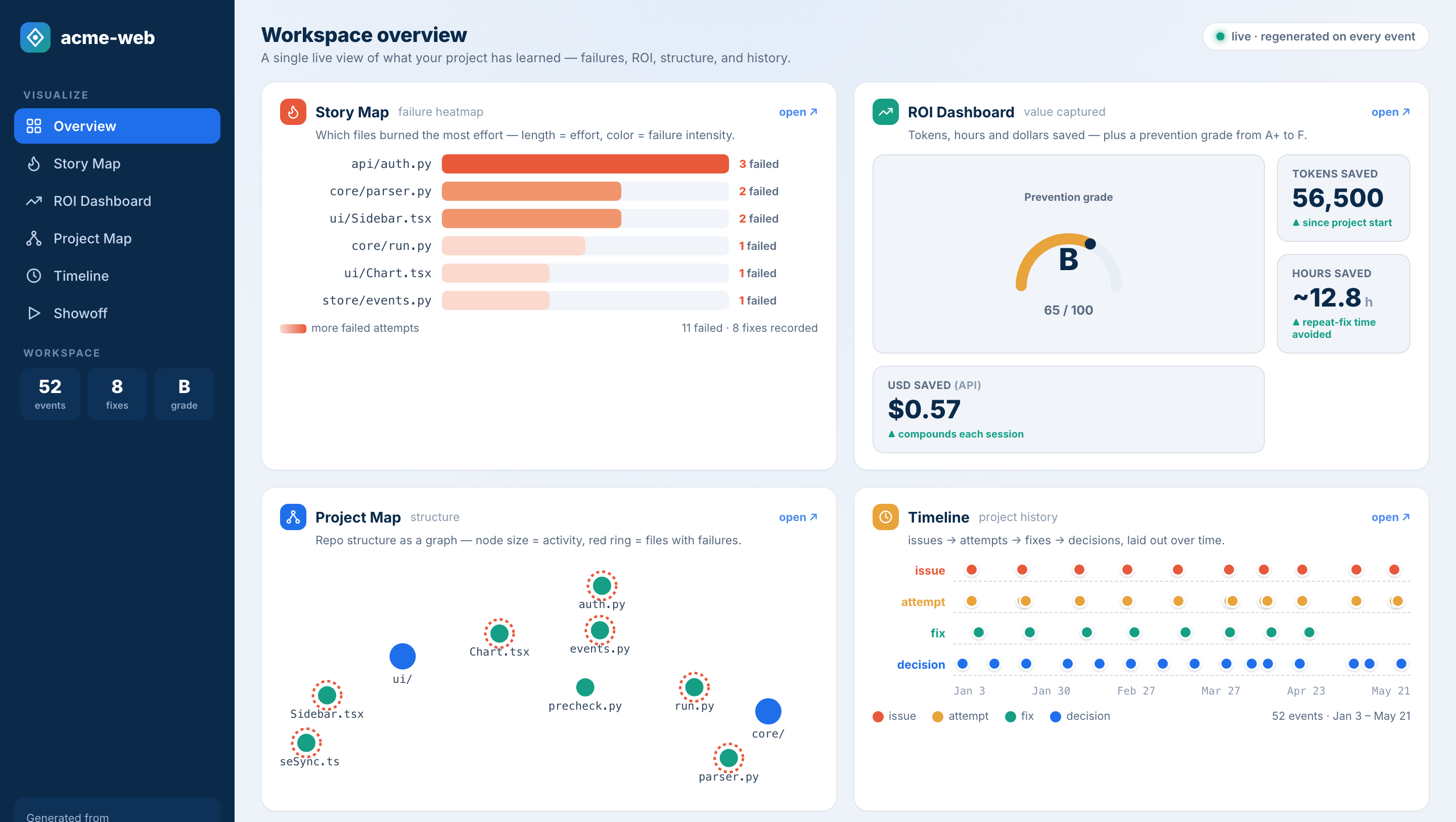}
\caption{The \texttt{pjm visualize} dashboard for a fictional project (\texttt{acme-web}, 52 events):
a failure heat map by file, the project's structure with files that have failed ringed in red, the
issue--attempt--fix--decision timeline, and the tool's own estimates of time and tokens saved. The page
is generated locally from \texttt{events.jsonl} and makes no network request.}
\label{fig:dashboard}
\end{figure}

\paragraph{The dashboard.} \texttt{pjm visualize} renders the same log as a local, self-contained HTML
page. Figure~\ref{fig:dashboard} shows it for a second fictional project with a longer history.

\bibliographystyle{plainnat}
\bibliography{references}

\end{document}